\documentclass{article}
\usepackage[utf8]{inputenc}

\usepackage{spconf}
\ninept

\usepackage{outlines}

\usepackage{graphics}
\usepackage{epstopdf}
\usepackage{graphicx}
\usepackage{subfigure}
\usepackage{latexsym}
\usepackage{amsmath,amsfonts,amssymb,mathdots}
\usepackage{comment}
\usepackage{float}

\usepackage{cite}
\usepackage{doi}

\usepackage{hyperref}
\hypersetup{
  colorlinks = true,
  allcolors = blue,
}
\usepackage[capitalise,nameinlink]{cleveref}

\usepackage{tikz}

\usepackage{xcolor}

\definecolor{myDarkBlue}{rgb}{0.0,0.0,0.3}
\definecolor{myBlue}{rgb}{0.0,0.0,1.0}
\definecolor{myWhite}{rgb}{1.0,1.0,1.0}
\definecolor{myRed}{rgb}{1.0,0.0,0.0}
\definecolor{myDarkRed}{rgb}{0.5,0.0,0.0}
\pgfdeclareverticalshading{seismic}{60bp}{color(0bp)=(myDarkBlue); color(15bp)=(myBlue); color(30bp)=(myWhite); color(45bp)=(myRed); color(60bp)=(myDarkRed)}

\usepackage{pgfplots}
\pgfplotsset{compat=1.17}
\usepgfplotslibrary{groupplots}
\usepgfplotslibrary{fillbetween}

\pgfplotscreateplotcyclelist{mycyclelist}{%
  {red, mark=o, mark options={solid}, solid, thick},
  {blue, mark=x, mark options={solid}, dashed, thick},
  {violet, mark=*, mark options={solid}, loosely dotted, thick},
  {orange, mark=o, mark options={solid}, dashed, thick},
  {black, mark=x, mark options={solid}, solid, thick}
}
\pgfplotsset{cycle list name=mycyclelist}

\DeclareFontEncoding{LS1}{}{}
\DeclareFontSubstitution{LS1}{stix}{m}{n}
\DeclareSymbolFont{symbols2}{LS1}{stixfrak} {m} {n}
\DeclareMathSymbol{\operp}{\mathbin}{symbols2}{"A8}
\DeclareMathOperator{\Ima}{Im}

\newcommand{\cev}[1]{\reflectbox{\ensuremath{\vec{\reflectbox{\ensuremath{#1}}}}}}

\title{Signal Processing on Cell Complexes}
\name{T.~Mitchell~Roddenberry$^+$, Michael~T.~Schaub$^\dagger$, Mustafa Hajij$^{*}$%
\thanks{%
TMR was partially supported by the 2020-21 K2I Exxon-Mobil Graduate Fellowship.
MTS acknowledges partial funding by the Ministry of Culture and Science (MKW) of the German State of North Rhine- Westphalia ("NRW Rückkehrprogramm") and  Excellence Strategy of the Federal Government and the Länder. MH was supported in part by the National Science Foundation (NSF, DMS-2134231).
Emails: \href{mitch@rice.edu}{mitch@rice.edu}, \href{schaub@cs.rwth-aachen.de}{schaub@cs.rwth-aachen.de}, \href{mhajij@scu.edu}{mhajij@scu.edu}.%
}
}
\address{
$^+$Department of Electrical and Computer Engineering, Rice University, USA \\
$^\dagger$Department of Computer Science, RWTH Aachen University, Germany\\ $^*$Department of Mathematics and Computer Science, Santa Clara University, USA }
\date{\today}

\begin{document}

\maketitle

\begin{abstract}
The processing of signals supported on non-Euclidean domains has attracted large interest recently.
Thus far, such non-Euclidean domains have been abstracted primarily as graphs with signals supported on the nodes, though the processing of signals on more general structures such as simplicial complexes has also been considered. 
In this paper, we give an introduction to signal processing on (abstract) regular cell complexes, which provide a unifying framework encompassing graphs, simplicial complexes, cubical complexes and various meshes as special cases. 
We discuss how appropriate Hodge Laplacians for these cell complexes can be derived. 
These Hodge Laplacians enable the construction of convolutional filters, which can be employed in linear filtering and non-linear filtering via neural networks defined on cell complexes. 
\end{abstract}


\section{Introduction}
Graph signal processing (GSP) has become important to study signals defined over non-Euclidean spaces.
GSP provides tools for processing signals supported on discrete spaces abstracted as graphs, leveraging ideas from both signal processing and graph theory~\cite{ShumanNarangFrossard2013}.
Recently, there has been increased interest in extending these ideas to more general topological spaces described, e.g., via simplicial complexes (SCs)~\cite{schaub2018denoising,barbarossa2016introduction,barbarossa2020topologicalmag,robinson2014topological,schaub2021signalbook,yang2021finite,barbarossa2020topological,schaub2021signal}.
Underpinning this development is the notion that encoding a richer set of relationships can a) help to capture relevant invariants in data supported on complex spaces more faithfully; b) enable the extraction of a richer set of structural features, ultimately leading to more powerful signal processing methods.

Similarly, geometric deep learning architectures, which have mostly been defined as neural networks on graphs, have recently been extended to domains modeled via SCs \cite{schaub2020random,ebli2020simplicial,roddenberry2021principled,bunch2020simplicial,hajij2021simplicial} and hypergraphs \cite{jiang2019dynamic,arya2018exploiting}.
In this context, having a richer set of interactions between neurons can reshape the information flow and enlarge the set of possible computations, which can increase the expressiveness of neural networks~\cite{bodnar2021weisfeiler}.
Similar ideas have also appeared in the study of dynamical systems defined on networks \cite{battiston2020networks}.

To some extent, these developments mirror ideas from the burgeoning area of topological data analysis (TDA)~\cite{edelsbrunner2010computational}.
The goal of topological data analysis is to understand the ``shape of data''~\cite{carlsson2009topology}, by constructing various complexes such as SCs, cubical complexes, etc., which provide discrete approximations of the (topological) space containing the observed data.
These complexes can then be analysed using tools such as persistent homology~\cite{edelsbrunner2008persistent}, or the mapper algorithm~\cite{singh2007topological}.
Interestingly, whereas in TDA many different kinds of constructions for complexes have been considered, outside of TDA most works have focused on SCs.

\begin{figure}[t!]
  \centering
   \includegraphics[scale=0.105]{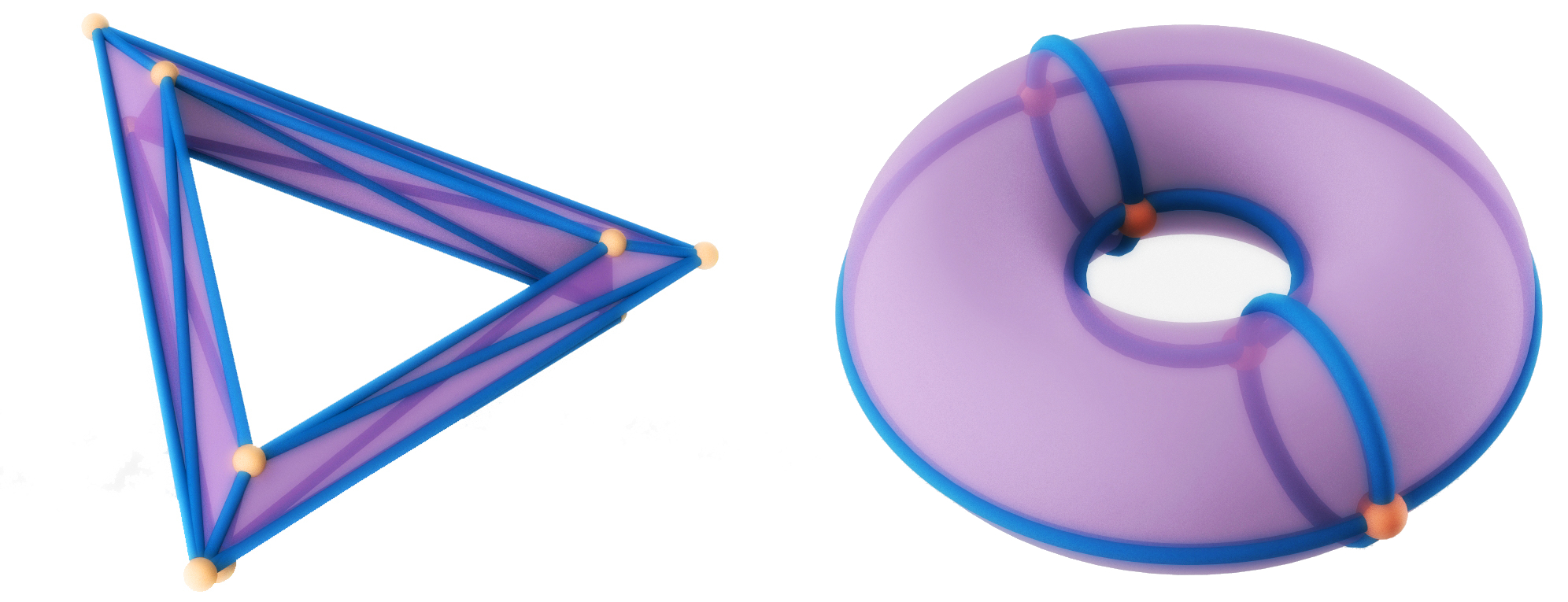}
    \caption{Compared to simplicial complexes, cell complexes can describe the same domain structure with fewer building blocks. For instance, the topology of a torus can be described much more succinctly by a cell complex (right) than a simplicial complex (left).}
  \label{fig:clique}
\end{figure}

Indeed, (abstract) SCs are arguably the simplest kind complexes and are already equipped with many desirable features: they extend graphs by allowing for relationships not just between vertex pairs, but also larger vertex sets. 
From a GSP point of view, SCs come equipped with natural shift operators in the form of Hodge Laplacians and their associated boundary maps \cite{schaub2020random,barbarossa2020topological,schaub2021signal}.
Despite these aspects, SCs have certain disadvantages as modelling tools for discrete domains.
In particular, within an SC any relationship between $k$ entities must be build from relationships of all the corresponding subsets of $k-1$ entities.
Consequently, encoding a space via an SC can require an relatively large amount of data and memory.
This complexity of encoding the domain can, e.g., impact the scalability of training neural networks; see~\cite{jia2020improving} for a discussion related to graphs.

Furthermore, if we want to process signals on an already discrete domain, e.g., traffic-flows on a (rectangular) street network, constructing a meaningful SC can be challenging due to an inherent lack of available simplices. 
To illustrate some of these shortcoming on an intuitive level, consider the example of a $2$-torus.
A compact way (but not the most compact) to describe the topology of the torus with an SC requires $9$ nodes, $27$ edges, and $18$ triangles.
This is shown in \cref{fig:clique}~(left), where the torus is triangulated to form an SC.

\noindent\textbf{Contributions.}
Can we find a simpler representation of this space that preserves the algebraic structure of SCs, which we can exploit for signal processing?
In this article, we argue that the theory of \emph{cell complexes} provides the necessary generalization for SCs to answer this question affirmatively.
Related ideas for signal processing on cell complexes can be found in~\cite{sardellitti2021topological,grady2010discrete}.
In particular, concepts for signal processing on general topological spaces (modeled, e.g., by cell compexes) have previously been discussed in~\cite{grady2010discrete,robinson2014topological}.

The purpose of this article is to present the main ingredients necessary for signal processing on cell complexes in an accessible and compact manner, generalizing the expositions in~\cite{schaub2020random,barbarossa2020topological,schaub2021signal} for simplicial complexes.
In particular, we provide a combinatorial characterization of cell complexes, which facilitates the construction of appropriate linear (shift) operators for signal processing.

At an intuitive level, a cell complex (CC) generalizes an SC in that rather than allowing only for simplices as building blocks, we are allowed to use cells of arbitrary shape. 
Accordingly, every cell can have an arbitrary number of lower dimensional cells on its boundary. 
This flexible type of complex yields, e.g., a substantially simpler model of the torus -- see \cref{fig:clique} (right).
Instead of having to divide the surface of the torus into $18$ triangles, the torus can be modeled with $4$ nodes, $8$ edges, and $4$ faces describing areas bounded by these edges.
Besides increased flexibility to model various domains, we may also gain numerical stability when replacing SCs with CCs.
For instance, in geometry processing, CCs in the form of polygonal meshes are preferable to SCs for solving PDEs on a surface, and are best suited for high-order surface modeling~\cite{bommes2013quad}. 

\noindent\textbf{Outline.} We provide mathematical background on CCs and their combinatorial  representations in \cref{sec:cell_complexes}.
We then introduce Hodge Laplacians on CCs and discuss how these matrices can serve as shift operators for signals supported on CCs.
With these Hodge Laplacians we construct linear convolutional filters for signals on CCs and illustrate their utility in \cref{sec:experiments}.
Finally, we show how to devise non-linear filters via neural networks defined on CCs in~\cref{sec:nonlinear}.
We close with a brief discussion on future work.

\section{Cell complexes and signals on cells}\label{sec:cell_complexes}

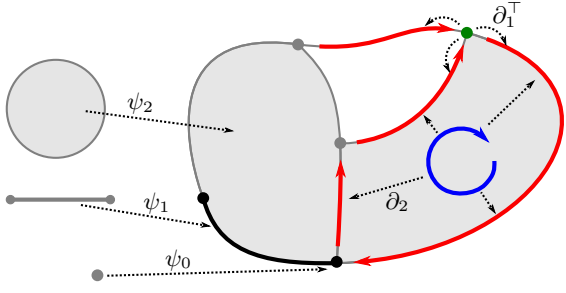
\begin{figure}[t!]
  \centering
\begingroup%
  \makeatletter%
  \providecommand\color[2][]{%
    \errmessage{(Inkscape) Color is used for the text in Inkscape, but the package 'color.sty' is not loaded}%
    \renewcommand\color[2][]{}%
  }%
  \providecommand\transparent[1]{%
    \errmessage{(Inkscape) Transparency is used (non-zero) for the text in Inkscape, but the package 'transparent.sty' is not loaded}%
    \renewcommand\transparent[1]{}%
  }%
  \providecommand\rotatebox[2]{#2}%
  \newcommand*\fsize{\dimexpr\f@size pt\relax}%
  \newcommand*\lineheight[1]{\fontsize{\fsize}{#1\fsize}\selectfont}%
  \ifx\svgwidth\undefined%
    \setlength{\unitlength}{210.38512502bp}%
    \ifx\svgscale\undefined%
      \relax%
    \else%
      \setlength{\unitlength}{\unitlength * \real{\svgscale}}%
    \fi%
  \else%
    \setlength{\unitlength}{\svgwidth}%
  \fi%
  \global\let\svgwidth\undefined%
  \global\let\svgscale\undefined%
  \makeatother%
  \begin{picture}(1,0.47556633)%
    \lineheight{1}%
    \setlength\tabcolsep{0pt}%
    \put(0,0){\includegraphics[width=\unitlength,page=1]{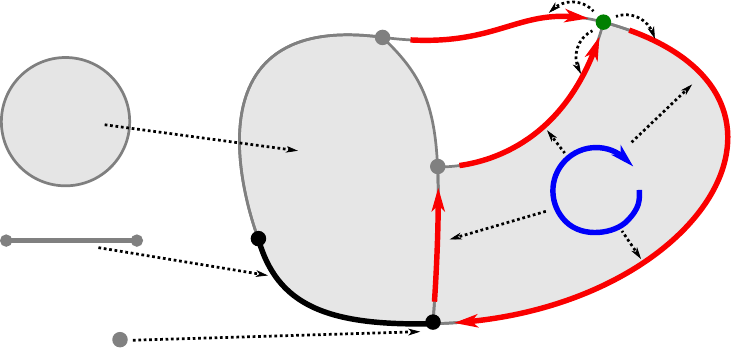}}%
    \put(0.21781852,0.30780574){\makebox(0,0)[lt]{\lineheight{1.25}\smash{\begin{tabular}[t]{l}$\psi_2$\end{tabular}}}}%
    \put(0.2468442,0.12920348){\makebox(0,0)[lt]{\lineheight{1.25}\smash{\begin{tabular}[t]{l}$\psi_1$\end{tabular}}}}%
    \put(0.28577694,0.02802985){\makebox(0,0)[lt]{\lineheight{1.25}\smash{\begin{tabular}[t]{l}$\psi_0$\end{tabular}}}}%
    \put(0.87201778,0.46229083){\makebox(0,0)[lt]{\lineheight{1.25}\smash{\begin{tabular}[t]{l}$\partial_1^\top$\end{tabular}}}}%
    \put(0.68023466,0.130){\makebox(0,0)[lt]{\lineheight{1.25}\smash{\begin{tabular}[t]{l}$\partial_2$\end{tabular}}}}%
  \end{picture}%
\endgroup%

    \caption{
    Illustration of a regular cell complex. Left: homeomorphisms $\psi_0,\psi_1,\psi_2$ to closed balls in Euclidean space. Right: cochains, chains, boundary maps, and coboundary maps.
    }
  \label{fig:cell-example}
\end{figure}

\noindent\textbf{Cell Complexes as topological domains.}
CCs form a natural generalization of almost all discrete domains that are important in practice, such as multigraphs, 3D shapes, SCs, point clouds, polyhedral complexes, delta complexes and cubical complexes \cite{hatcher2005algebraic}.
In this paper, all CCs are assumed to be regular and consist of finitely many cells. 

It is instructive to informally think of CCs as geometric spaces inductively built up from \textit{cells}.
Starting with points as $0$-cells, $k$-cells are defined such that their geometric boundary consists of a (finite) collection of $k-1$ cells.
For instance, a line is a $1$-cell with two $0$ dimensional points as boundary, and a polygon may serve as a $2$-cell, with a boundary consisting of a set of line segments.
We may now think of a CC as a topological space consisting of a set of cells for which we have ``glued together'' the boundaries of a finite collection of cells, such that two cells may now have a joint boundary.
Furthermore, by definition, the boundaries of every cell are always part of the CC.
Thus, a CC typically consists of a collection of cells of varying dimensions which are related via their boundaries.
In particular, the cells that bound a cell are called the \emph{faces} of that cell.
The faces of a $k$-cell in a regular cell complex are $(k-1)$-cells: the set of faces of a $0$-cell is the empty set.

Formally, a regular cell complex $\mathcal X$ is a topological space with a partition into subspaces (cells) $\{X_\alpha\}$, such that for all $\alpha,\beta$, we have~\cite{hatcher2005algebraic,hansen2019toward}:
a) $\overline{X}_\alpha\cap X_\beta \neq \emptyset$ only if $X_\beta \subset \overline{X}_\alpha$, where $\overline{X}$ refers to the closure of set $X$;
b) there exists a homeomorphism $\psi$ (a continuous bijection with a continuous inverse) from $X_\alpha$ to $\mathbb{R}^{n_\alpha}$ for some $n_\alpha\in \mathbb{N}$, called the dimension of the cell;
c) there exists a homeomorphism, called an attaching map, of the closed unit ball 
to $\overline{X}_\alpha$ that homeomorphically maps the interior of the ball 
onto $X_\alpha$.
\cref{fig:cell-example} depicts these definitions.
In particular, each $2$-cell is bounded by $1$-cells (edges), and each $1$-cell is bounded by $0$-cells (nodes).
Moreover, the cells are homeomorphic to closed balls of corresponding dimension in Euclidean space: the closed disc, the closed interval, and a point for $2$-cells, $1$-cells, and $0$-cells, respectively.

For simplicity, we will concentrate on CCs with at most $2$ dimensional cells in the following, even though the theory extends beyond this case. 
As explained in the following subsections, we can describe CC in a compact combinatorial manner, akin to graphs.

\noindent\textbf{A combinatorial description of cell complexes.}
Consider a cell complex $\mathcal X$ and denote the set of $k$-cells in $\mathcal X$ by $X^k$, with cardinality $N_k=|X^k|$.
In analogy to graphs, we call the elements of the set $X^0$ the nodes and denote them by $v_i$ for $i\in 1,\ldots, N_0$.
Similarly, we call the elements of $X^1$ the edges of the cell complex and denote them by $e_i$ for $i \in 1,\ldots, N_1$.
Finally, we call the elements of $X^2$ the $2$-cells of the complex, and denote them by $\theta_i$ for $i \in 1\ldots, N_2$.

It is customary to introduce a reference orientation to each edge and 2-cell.
This choice of an orientation is a matter of bookkeeping: similar to labeling the nodes of a graph to encode it in an adjacency matrix, those reference orientations are essential for performing appropriate numerical computations with (flow) signals on cell complexes.
We denote the $k$-th oriented edge going from node $v_i$ to node $v_j$ via the ordered tuple $\Vec{e}_k = [v_i, v_j]$.
The corresponding oppositely oriented edge will be denoted by $\cev{e}_k = [v_j, v_i]$.
The $k$-th oriented $2$-cell will be denoted by the ordered tuple $\Vec{\theta}_k = [\Vec{e}_{i}, \ldots, \Vec{e}_{j}]$ defined by a sequence $\vec{e}_i,\ldots,\vec{e}_j$ of oriented edges that form a non-intersecting closed path of length $m \geq 2$.
Note that the path defining $\Vec{\theta}_k$ may contain edges $\cev e_\ell$ traversed against their reference orientation as well.
Any cyclic permutation of the ordered tuple $\Vec{\theta}_k$ defines the same $2$-cell; a flip of both the orientation and ordering of all the edges defining $\vec{\theta}_k$ corresponds to a change in the orientation of the $2$-cell, i.e., $\cev{\theta}_k = [\cev{e}_{j}, \ldots, \cev{e}_{i}]$.
Unlike for SC just flipping a single edge in $\vec{\theta}_k$ may lead to an inconsistent closed path (cell), and thus is not feasible, in general.
Indeed, in an SC any permutation of the vertex labels of a simplex leads to consistent path.

Given a reference orientation for each cell, for each $k$ we can define a finite-dimensional vector space $\mathcal{C}_k$ with coefficients in $\mathbb{R}$ whose basis elements are the oriented $k$-cells. 
An element $\chi_k \in \mathcal{C}_k$ is called a $k$-\emph{chain} and may be thought of as a formal linear combination of these basis elements.
For instance, a $1$-chain may be written as $\chi_1 = \sum_i a_i \Vec{e}_i$ for some $a_i\in\mathbb{R}$.
We augment the construction of $\mathcal{C}_k$ by one more aspect: an orientation change of the basis elements is defined to correspond to a change in the sign of the coefficient $a_i$.
Hence, if we flip the orientation of a basis element, we have to multiply the corresponding coefficient $a_i$ by  $-1$, e.g., $a_1\vec{e}_1 = -a_1\cev{e}_1$.

Observe that for any $k$ the space $\mathcal{C}_k$ is isomorphic to $\mathbb{R}^{N_k}$, so we may compactly represent each element $\chi_k \in \mathcal{C}_k$ by a vector ${c=(a_1,...,a_{N_k})^\top}$.
Finally, we endow each space $\mathcal C_k$ with the standard $\ell_2$ inner product $\langle c_1, c_2\rangle  = c^\top_1 c_2$, and thus give $\mathcal{C}_k$ the structure of a finite-dimensional Hilbert space.

\noindent\textbf{Signals on cell complexes: cochains.}
In the following, we concentrate on edge-signals on CCs, which we will think of as flows.
Such flows can be conveniently described by so-called cochains, which may be thought of as maps that assign a scalar value to each cell.
Such a cochain may be interpreted as a signal supported on the cells of the complex and be conveniently represented by a vector.

Formally, the space of \emph{$k$-cochains} is the dual space of the space of $k$-chains and denoted as $\mathcal{C}^k:=\mathcal{C}_k^*$.
In the finite case, these spaces are isomorphic, but interpreted in different ways.
Namely, $k$-chains are interpreted as weighted collections of cells, while $k$-cochains are functions assigning scalar values to $k$-cells. 
In the context of smooth manifolds, $k$-cochains are related to differential forms, which one integrates over a domain described by a $k$-chain.

\noindent\textbf{Mappings between different dimensions: boundary maps.}
Chains of different dimensions can be related via boundary maps $\partial_k: \mathcal{C}_k\rightarrow \mathcal{C}_{k-1}$, which map a chain to a sum of its boundary components.
We can define these linear maps via their action on the basis elements, which are simply each of the oriented cells.
In our setting, we have $\partial_1(\vec{e}) = \partial([v_i,v_j]) = v_i - v_j$ and $\partial_2(\vec{\theta}) = \partial_2([\vec{e}_{i_1}, \ldots,\vec{e}_{i_m}]) = \sum_{j=1}^m \vec{e}_{i_j}$.
Since all the spaces involved are finite dimensional we can represent these boundary maps via matrices $\mathbf{B}_1$ and $\mathbf{B}_2$, respectively, which act on the corresponding vector representations of the chains.

The dual of these boundary maps are the co-boundary maps ${\partial_k^\top: \mathcal{C}^{k-1} \rightarrow \mathcal{C}^{k}}$, which map cochains of lower to higher-dimensions.
Given the inner-product structure of $\mathcal{C}_k$ defined above, these are simply the adjoint maps to $\partial_k$ and their matrix representation is accordingly given by $\mathbf{B}_1^\top$ and $\mathbf{B}_2^\top$, respectively.
\cref{fig:cell-example} illustrates all of these concepts.
We illustrate a $2$-chain using an oriented circular arrow on the rightmost face.
Applying the boundary map $\partial_2$ to this $2$-chain yields a sum of the oriented edges bounding the $2$-cell, illustrated by the red arrows.
Similarly, if we consider a $0$-cochain, which simply assigns a real number to each vertex as illustrated by the green vertex, the coboundary operator applied to that vertex induces a nonzero $1$-cochain on the edges incident to that node, having no effect elsewhere in the CC.

\section{Linear filters on cell complexes and the Hodge Laplacian}\label{sec:hodge}

Having defined the domain (regular cell complexes) and the signals of interest (cochains), we now consider how we can use the structure of the domain to process said signals.
We take a GSP approach to this problem.
First, we establish a suitable operator that reflects the combinatorial structure of the domain.
Then, under the hypothesis that the signal's properties are reflective of the domain, we use the chosen operator to build filters that reinforce the hypothesis.

Given a regular CC $\mathcal X$ with boundary matrices as defined above we define the $k$-th combinatorial Hodge Laplacian by :
\begin{equation}
    \mathbf{L}_k = \mathbf{B}_k^\top \mathbf{B}_k + \mathbf{B}_{k+1}\mathbf{B}_{k+1}^\top
\end{equation}
The $k$-th Hodge Laplacian provides a mapping from the space of signals on $k$-cells onto itself.
Specifically, the $0$-th Hodge Laplacian operator, is simply the graph Laplacian $\mathbf L_0=\mathbf{B}_1\mathbf{B}_1^\top  $ of the graph corresponding to the $1$-skeleton of the CC (note that $\mathbf{B}_0:=0$ by convention).
Similar to the graph Laplacian, the Hodge Laplacian is a positive semidefinite operator and its eigenvalues can be interpreted in terms of non-negative frequencies~\cite{schaub2021signal,barbarossa2020topological}.

Using the fact that $\partial_k \circ \partial_{k+1}=0$ and the definition of $L_k$, it can be shown that the space of $k$-cochains on $\mathcal X$ admits a so-called Hodge-decomposition \cite{lim2020hodge,schaub2020random,grady2010discrete}:
\begin{equation}
    \mathcal{C}^k= \Ima (\partial_{k+1}) \oplus \Ima (\partial_{k}^\top) \oplus \ker(L_k).
\end{equation}
In the context of flow-signals, i.e., $1$-cochains, this decomposition is the discrete equivalent of the celebrated Helmholtz decomposition for a continuous vector field into curl, gradient and harmonic components.
Specifically, we can create any gradient signal via a vector $\phi$ assigning a potential $\phi_i$ to each vertex $i$ in the complex, and then applying the co-boundary map $\partial_1^\top$.
Likewise, any curl flow can be created by applying the boundary map $\partial_2$ to a vector $\eta$ of $2$-cell potentials.
(Note how this is precisely analogous to the scalar and vector potentials appearing in the Helmholtz decomposition).

Importantly, it can be shown that each of the above discussed three subspaces is spanned by a set of eigenvectors of the Hodge Laplacian.
Namely, the eigenvectors of the \emph{lower Laplacian} $\mathbf L_k^{low} = \mathbf{B}_k^\top \mathbf{B}_k$ precisely span $\Ima (\mathbf{B}^\top_{k})$ (the gradient space); the eigenvectors of the \emph{upper Laplacian} $\mathbf L_k^{up} = \mathbf{B}_{k+1}\mathbf{B}_{k+1}^\top$ span $\Ima (\mathbf{B}_{k+1})$ (curl space), and the eigenvectors associated to zero eigenvalues span the harmonic subspace.

Using the Hodge Laplacian in lieu of the graph Laplacian as a shift operator for signals supported on CCs, we can define a \emph{cell-filter} $\mathbf H$ of order $M$ via the matrix polynomial
\begin{equation}\label{eq:filter-polynomial}
    \mathbf{H}(\mathbf{L}_k) = \sum_{\ell=0}^L h_\ell \mathbf{L}_k^\ell.
\end{equation}
Examining \eqref{eq:filter-polynomial}, we see that $\mathbf{H}(\mathbf{L}_k)$ is simply a weighted sum of powers of $\mathbf{L}_k$.
Considering the way that $\mathbf{L}_k$ is defined, one can check that $[\mathbf{L}_k]_{ij}\neq 0$ only if the $i^{\mathrm{th}}$ and $j^{\mathrm{th}}$ $k$-cells are incident to one another.
Then, it follows that for $\ell\geq 0$, we have that $[\mathbf{L}_k^\ell]_{ij}\neq 0$ only if the $i^{\mathrm{th}}$ and $j^{\mathrm{th}}$ $k$-cells are at most $\ell$ ``hops'' away from one another in the CC.
Thus, higher order terms in a polynomial reflect increasingly longer-range relationships among elements in the CC.

Much like in GSP, a typical hypothesis is that signals on the domain are smooth with respect to the Laplace operator.
For instance,  one might assume that $1$-cochains (edge flows) are approximately harmonic~\cite{schaub2018denoising,jia2019ssl}, and are thus mostly supported by the eigenvectors of the Hodge Laplacian with small eigenvalues.
To denoise or represent such a signal, then, it may be appropriate to design the polynomial $\mathbf{H}(\cdot)$ as a \emph{low-pass filter}.
That is to say, $\mathbf{H}(\lambda)$ should take values close to one for small $\lambda$, and be close to zero for large $\lambda$.

\section{Numerical Illustration}\label{sec:experiments}

\begin{figure}
  \centering
  \resizebox{\linewidth}{!}{\begin{tikzpicture}[shading=seismic]
  \tikzstyle{every node}=[font=\Large]
  \pgfplotstableset{col sep=comma}
  \begin{groupplot}[
    group style={
      group size=2 by 1,
      group name=myplots,
      horizontal sep=6em,
    },
    ]
    
    \nextgroupplot[hide axis, enlargelimits=false,]
    \addplot graphics[xmin=0,xmax=1,ymin=0,ymax=1] {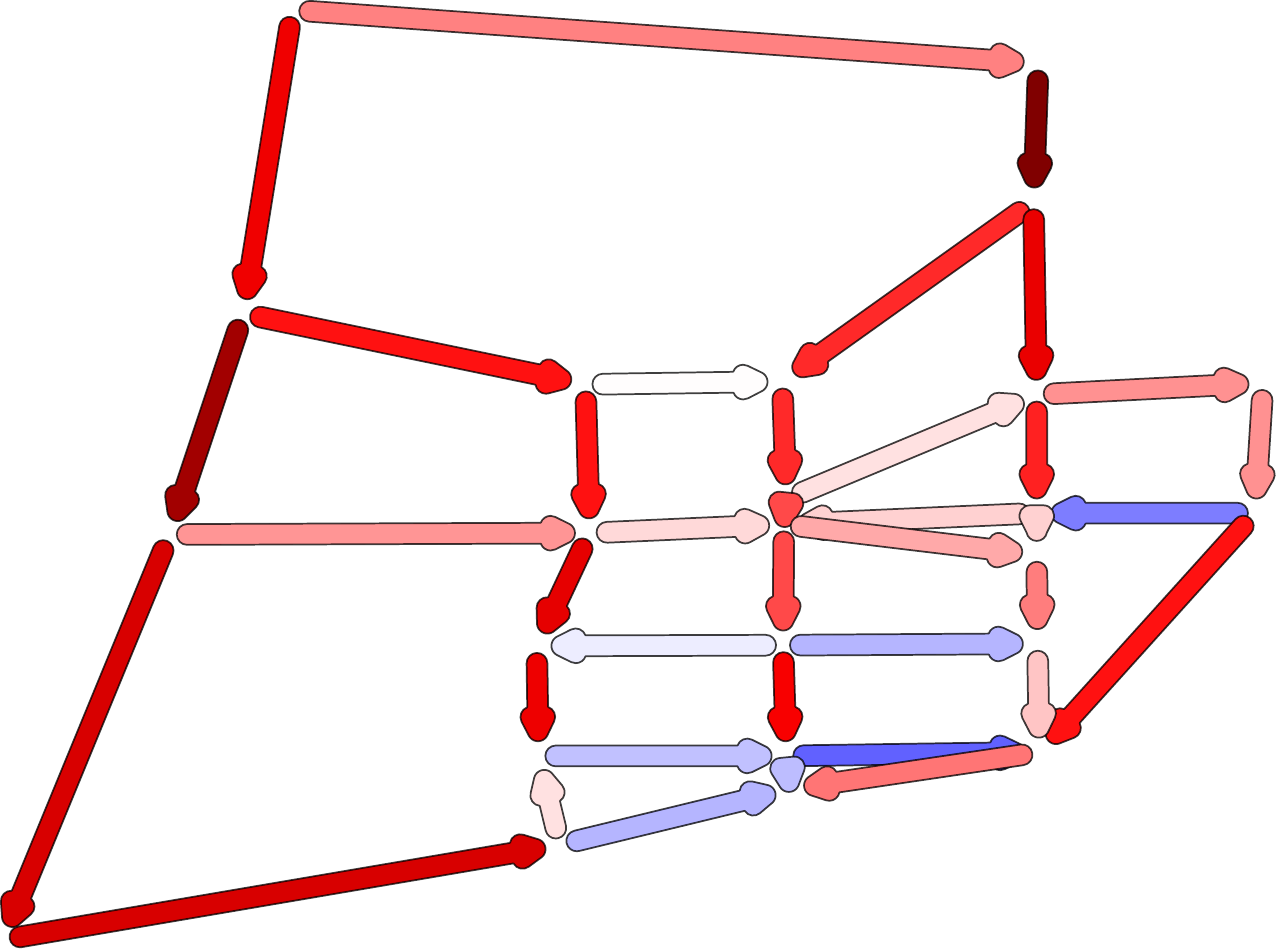};

    \nextgroupplot[
    xlabel={Noise St. Dev.},
    ylabel={MSE},
    xmode=log,
    legend columns=2,
    legend style={
        at={(0.03,0.7)},
        anchor=north west,
    },
    ylabel near ticks,
    xmajorgrids,
    ymajorgrids,
    mark size=2,
    ]
    
    \addplot+ table[x=power, y=std] {filter_mse.csv};
    \addlegendentry{Cells}
    \addplot+ table[x=power, y=sc] {filter_mse.csv};
    \addlegendentry{Simp.}
    \addplot+ table[x=power, y=down] {filter_mse.csv};
    \addlegendentry{Edge}
    \addplot+ table[x=power, y=lg] {filter_mse.csv};
    \addlegendentry{LG}
    
  \end{groupplot}
  
  \foreach \plt/\lab in {c1r1/a,c2r1/b} {
    \node[anchor=north west,fill=white] at (myplots \plt.north west) {(\lab)};
  }
  
  \shadedraw[color=black, very thin, shading angle=-90] ($(myplots c1r1.south west)+(0.5,-0.25)$) rectangle ($(myplots c1r1.south east)+(-0.5,-0.75)$);
  \node[anchor=north west] at ($(myplots c1r1.south west)+(0.5,-0.75)$) {$-1.0$};
  \node[anchor=north] at ($(myplots c1r1.south west)!0.5!(myplots c1r1.south east)+(0,-0.75)$) {$0.0$};
  \node[anchor=north east] at ($(myplots c1r1.south east)+(-0.5,-0.75)$) {$+1.0$};
\end{tikzpicture}}
  \caption{%
    Denoising flows with cellular Hodge Laplacians.
    (a) Synthetic flow over Sioux Falls road network.
    Each intersection is a node, each road is an edge, and each planar area is a face in the constructed CC.
    Orientations of edges are indicated by arrows, and color encodes flow volume relative to the orientation.
    (b) Mean squared error in flow denoising task under additive white Gaussian noise.
  }
  \label{fig:filtering-results}
\end{figure}
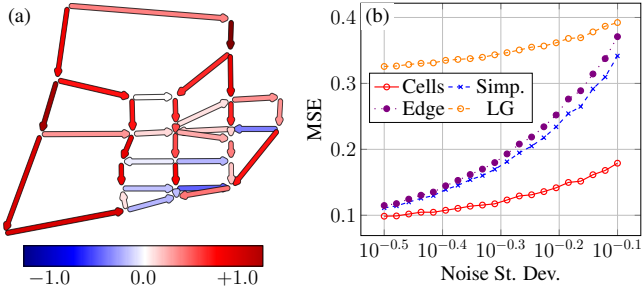

To illustrate the utility of the Hodge Laplacian as a shift operator for filtering, we consider a flow denoising task.
We consider flows ($1$-cochains) on the Sioux Falls road network~\cite{leblanc1975algorithm}, as illustrated in \cref{fig:filtering-results}~(a).
The flow corresponds to the sum of $1000$ trajectories on the edges created by picking a source node from the northern side, a target node from the southern side, and then generating a random walk that starts at the source and terminates at the target.

As the Sioux Falls road network forms a planar graph, we take each ``hole'' in the graph to be a $2$-cell, turning the network into a regular CC with $(24,38,15)$ $0$-cells, $1$-cells, and $2$-cells, respectively.
Given the initial flow $\mathbf{f}^*$, created as described above, we assume that we observe a noisy version of this flow, of the form $\mathbf{f} = \mathbf{f}^* + \boldsymbol{\epsilon}$, where $\boldsymbol{\epsilon}$ is white Gaussian noise.
To estimate the true flow $\mathbf{f}^*$, we perform the computation $\hat{\mathbf{f}} = \mathbf{H}\mathbf{f}$ via a filter $\mathbf{H}$ defined as follows:
\begin{equation}
    \mathbf{H}(\mathbf{S}) = \left(\mathbf I-\lambda_{\max}^{-1}{\mathbf{S}}{}\right)^3,
\end{equation}
where $\lambda_{\max}$ is the maximum eigenvalue of the matrix $\mathbf S$.
Note that the described filter is a low-pass filter, in the sense that it removes components of signals that correspond to the eigenvectors of the matrix $\mathbf{S}$ with large eigenvalues.
We apply this filter for multiple operators $\mathbf S:\mathcal{C}^1\to\mathcal{C}^1$, including
a) the cellular Hodge Laplacian,
b) the simplicial Hodge Laplacian, which only uses faces bounded by exactly three edges,
c) the edge Laplacian~\cite{schaub2018denoising}, and
d) the linegraph Laplacian~\cite{schaub2018denoising}.
We report the mean squared error over 500 trials in recovering the ground truth flow signal in \cref{fig:filtering-results}~(b).
One immediately sees that the linegraph Laplacian is poorly suited to the flow denoising task, for reasons explained by~\cite{schaub2018denoising,schaub2021signal}.
The simplicial and edge Laplacians also fare poorly, particularly under large amounts of noise, due to their failure to leverage the relevant cellular structure of the network.
Finally, the cellular Laplacian performs the best, since it captures more general phenomena by expressing the geometry of the underlying CC.
For instance, the notion of curl in a vector field can only be measured with respect to $2$-cells: if there is a $2$-cell that (without splitting the cell artificially into simplices) can only be represented by a CC and not a simplicial complex, there is not an immediate way to define a notion of curl around that cell using a simplicial complex model.
The good performance of this operator indicates that the flexibility of CC for modeling flow signals has utility over the stricter class of simplicial complexes, even in relatively benign examples such as this one.

\section{Nonlinear Filters : Deep Learning on Cell Complexes}\label{sec:nonlinear}
After we have seen how to construct linear filters for cell-signals in terms of polynomials of the Hodge Laplacian, in this section we discuss how these filters can serve as building blocks for more general nonlinear filters: neural networks supported on CCs \cite{hajijcell}.

\noindent\textbf{Convolutional Neural Networks for $k$-chains.}
The success of deep convolutional networks for images~\cite{krizhevsky2012imagenet} has been a main driver for the extension of such models to non-Euclidean domains, primarily modeled as graphs~\cite{kipf2016semi}.
In their simplest form, convolutional graph neural networks consist of layers that each comprise a linear graph filter based on 
the graph Laplacian or an appropriate adjacency matrix as shift operator, and a pointwise nonlinearity.
In the following we use $\mathbf{s}_k^j$ to denote the vector representation of $k$-cochains (signals) at the $j$-th layer of a neural network.

Building on previous works that exploited the 1-Hodge Laplacian operator for linear filtering (e.g., \cite{schaub2018denoising,barbarossa2020topological,barbarossa2020topologicalmag,schaub2021signal}), some of the earliest work in this direction was done in~\cite{roddenberry2019hodgenet}. 
Namely, letting $\mathbf{s}_1^0\in~\mathbb{R}^{{N_1}\times F } $ denote the vector of initial edge-signals, where $F$ is the dimension of the feature space, \cite{roddenberry2019hodgenet} considered neural network layers that basically have the form:
\begin{equation}
    \mathbf{s}^{k+1}_1=\sigma( \mathbf{H} \mathbf{s}^{k}_1 \mathbf{W}^{k+1} ),
\end{equation}
where $\mathbf{H}$ is a polynomial of the Hodge Laplacian $\mathbf{L}_1$, and $\mathbf{W}^{k+1}\in\mathbb{R}^{F\times F}$ is a learnable parameter matrix. 
This setting was extended to other dimensions for simplicial~\cite{ebli2020simplicial} and cell complexes~\cite{hajijcell}:
\begin{equation}
\label{conv_simple}
    \mathbf{s}^{k+1}_l=\sigma( \mathbf{H} \mathbf{s}^{k}_l \mathbf{W}^{k+1} ),
\end{equation}
where $\mathbf{H}$ is a polynomial of the Hodge Laplacian $\mathbf{L}_k$ or any matrix capturing the (upper and lower) adjacency information of the $k$-cells.

\noindent\textbf{Separate weightings for up and down Laplacian.}
As discussed in~\cref{sec:hodge}, the Hodge decomposition allows us to decompose an arbitrary signal into gradient, curl and harmonic parts, described by the upper and lower parts of the Hodge Laplacian.
This suggests that separate weights for upper and lower Laplacian are useful to enable a separate tuning of the curl and gradient spaces, as demonstrated in~\cite{yang2021finite}. 
This idea has also been recently exploited in the neural network architecture proposed by~\cite{roddenberry2021principled}, defined via: 
\begin{equation}
\label{mitch_principled}
\mathbf{s}^{l+1}_{1}\!\!=\sigma(\mathbf{L}_1^{low} \mathbf{s}_1^l \mathbf{W}_0^l + \mathbf{s}_1^l \mathbf{W}_1^l + \mathbf{L}_1^{up} \mathbf{s}_1^l \mathbf{W}_2^l)\!=:\!\sigma (\Phi(\mathbf{s}_1^l,\mathbf{W}_i^l)),
\end{equation}
where $\mathbf{W}_i^l$ are again parameter matrices.
Such a separation of weights was used, e.g., by~\cite{roddenberry2021principled} for trajectory projection tasks.

\noindent\textbf{Neural networks for cochains of different dimensions.} 
The architectures suggested above can be extended to simultaneously process cochains of different dimensions. Namely, thinking of the Hodge Laplacian as an operator that maps a $k-$cochain to another $k$-cochain, equations (\ref{conv_simple},\ref{mitch_principled}) can be extended by replacing the Hodge Laplacian by an appropriate linear operators $\mathbf{G}_i : \mathcal{C}^i\to \mathcal{C}^j$, such as (co)boundary maps) that map upper or lower co-chains to co-chains according to their incidence relations. 
This yields the general form:  
\begin{equation}
    \mathbf{s}_j^{l+1} =\sigma\left (\Phi(\mathbf{s}_j^l,\mathbf{W}_i^l) + \sum_i \mathbf{G}_i \mathbf{s}_i^{l} \mathbf{\tilde{W}}_i^{k+1} \right),
\end{equation}
with parameter matrices $\mathbf{\tilde{W}}_i^j$ and $\mathbf{W}_i^j$.
Such variants have appeared in multiple works recently including \cite{bunch2020simplicial,hajij2021simplicial, schaub2020random, schaub2021signal, roddenberry2021principled,hajijcell}. We remark that we have only scratched the surface here, and there remains a large space of architectures to be explored and analysed.

\section{Discussion}

Based on the recent work in signal processing on simplicial complexes, we have considered definitions of signals and boundary operators on regular cell complexes.
In doing so, we allow for more flexible models of non-Euclidean spaces that allow for implementation of familiar filtering operations, simply by taking polynomials of suitably defined Hodge Laplacians.
By reducing the typical topological definition of a regular cell complex down to a purely combinatorial description, we demonstrate how these operators can be treated as simple matrices acting on real vectors.
The flexibility afforded by modeling spaces as cell complexes proves advantageous, as it is not restricted by the simplicial structure of existing approaches.
Moreover, these simple operators form the foundation of modern graph neural network architectures, which interleave linear filtering operations with nonlinear activation functions.

\newpage
\bibliographystyle{ieeetr}
\bibliography{refs}

\begin{thebibliography}{10}

\bibitem{ShumanNarangFrossard2013}
D.~I. Shuman, S.~K. Narang, P.~Frossard, A.~Ortega, and P.~Vandergheynst, ``The
  emerging field of signal processing on graphs: Extending high-dimensional
  data analysis to networks and other irregular domains,'' {\em Signal
  Processing Magazine}, vol.~30, no.~3, pp.~83--98, 2013.

\bibitem{schaub2018denoising}
M.~T. Schaub and S.~Segarra, ``Flow smoothing and denoising: Graph signal
  processing in the edge-space,'' in {\em Global Conference on Signal and
  Information Processing}, pp.~735--739, 2018.

\bibitem{barbarossa2016introduction}
S.~Barbarossa and M.~Tsitsvero, ``An introduction to hypergraph signal
  processing,'' in {\em International Conference on Acoustics, Speech, and
  Signal Processing}, pp.~6425--6429, 2016.

\bibitem{barbarossa2020topologicalmag}
S.~Barbarossa and S.~Sardellitti, ``Topological signal processing: Making sense
  of data building on multiway relations,'' {\em Signal Processing Magazine},
  vol.~37, no.~6, pp.~174--183, 2020.

\bibitem{robinson2014topological}
M.~Robinson, {\em Topological Signal Processing}.
\newblock Springer, 2014.

\bibitem{schaub2021signalbook}
M.~T. Schaub, J.-B. Seby, F.~Frantzen, T.~M. Roddenberry, Y.~Zhu, and
  S.~Segarra, ``Signal processing on simplicial complexes,'' {\em arXiv
  preprint arXiv:2106.07471}, 2021.

\bibitem{yang2021finite}
M.~Yang, E.~Isufi, M.~T. Schaub, and G.~Leus, ``Finite impulse response filters
  for simplicial complexes,'' {\em European Signal Processing Conference},
  2021.

\bibitem{barbarossa2020topological}
S.~Barbarossa and S.~Sardellitti, ``Topological signal processing over
  simplicial complexes,'' {\em Transactions on Signal Processing}, vol.~68,
  pp.~2992--3007, 2020.

\bibitem{schaub2021signal}
M.~T. Schaub, Y.~Zhu, J.-B. Seby, T.~M. Roddenberry, and S.~Segarra, ``Signal
  processing on higher-order networks: Livin’on the edge... and beyond,''
  {\em Signal Processing}, vol.~187, p.~108149, 2021.

\bibitem{schaub2020random}
M.~T. Schaub, A.~R. Benson, P.~Horn, G.~Lippner, and A.~Jadbabaie, ``{Random
  walks on simplicial complexes and the normalized Hodge 1-Laplacian},'' {\em
  SIAM Review}, vol.~62, no.~2, pp.~353--391, 2020.

\bibitem{ebli2020simplicial}
S.~Ebli, M.~Defferrard, and G.~Spreemann, ``Simplicial neural networks,'' {\em
  NeurIPS Workshop on TDA and Beyond}, 2020.

\bibitem{roddenberry2021principled}
T.~M. Roddenberry, N.~Glaze, and S.~Segarra, ``Principled simplicial neural
  networks for trajectory prediction,'' in {\em International Conference on
  Machine Learning}, pp.~9020--9029, 2021.

\bibitem{bunch2020simplicial}
E.~Bunch, Q.~You, G.~Fung, and V.~Singh, ``Simplicial 2-complex convolutional
  neural nets,'' {\em NeurIPS Workshop on TDA and Beyond}, 2020.

\bibitem{hajij2021simplicial}
M.~Hajij, T.~Papamarkou, V.~Maroulas, G.~Zamzmi, and X.~Cai, ``Simplicial
  complex representation learning,'' {\em Machine Learning on Graphs Workshop
  at WSDM’22}, 2022.

\bibitem{jiang2019dynamic}
J.~Jiang, Y.~Wei, Y.~Feng, J.~Cao, and Y.~Gao, ``Dynamic hypergraph neural
  networks.,'' in {\em International Joint Conference on Artificial
  Intelligence}, pp.~2635--2641, 2019.

\bibitem{arya2018exploiting}
D.~Arya and M.~Worring, ``Exploiting relational information in social networks
  using geometric deep learning on hypergraphs,'' in {\em ACM International
  Conference on Multimedia Retrieval}, pp.~117--125, 2018.

\bibitem{bodnar2021weisfeiler}
C.~Bodnar, F.~Frasca, Y.~Wang, N.~Otter, G.~F. Montufar, P.~Lio, and
  M.~Bronstein, ``Weisfeiler and lehman go topological: Message passing
  simplicial networks,'' in {\em International Conference on Machine Learning},
  pp.~1026--1037, PMLR, 2021.

\bibitem{battiston2020networks}
F.~Battiston, G.~Cencetti, I.~Iacopini, V.~Latora, M.~Lucas, A.~Patania, J.-G.
  Young, and G.~Petri, ``Networks beyond pairwise interactions: Structure and
  dynamics,'' {\em Physics Reports}, vol.~874, pp.~1--92, 2020.

\bibitem{edelsbrunner2010computational}
H.~Edelsbrunner and J.~Harer, {\em Computational topology: an introduction}.
\newblock American Mathematical Society, 2010.

\bibitem{carlsson2009topology}
G.~Carlsson, ``Topology and data,'' {\em Bulletin of the American Mathematical
  Society}, vol.~46, no.~2, pp.~255--308, 2009.

\bibitem{edelsbrunner2008persistent}
H.~Edelsbrunner, J.~Harer, {\em et~al.}, ``Persistent homology: A survey,''
  {\em Contemporary Mathematics}, vol.~453, pp.~257--282, 2008.

\bibitem{singh2007topological}
G.~Singh, F.~M{\'e}moli, G.~E. Carlsson, {\em et~al.}, ``Topological methods
  for the analysis of high dimensional data sets and 3d object recognition.,''
  {\em PBG @ Eurographics}, vol.~2, 2007.

\bibitem{jia2020improving}
Z.~Jia, S.~Lin, M.~Gao, M.~Zaharia, and A.~Aiken, ``Improving the accuracy,
  scalability, and performance of graph neural networks with {ROC},'' {\em
  Proceedings of Machine Learning and Systems}, pp.~187--198, 2020.

\bibitem{sardellitti2021topological}
S.~Sardellitti, S.~Barbarossa, and L.~Testa, ``Topological signal processing
  over cell complexes,'' in {\em Asilomar Conference on Signals, Systems, and
  Computers}, 2021.

\bibitem{grady2010discrete}
L.~J. Grady and J.~R. Polimeni, {\em Discrete Calculus}.
\newblock Springer, 2010.

\bibitem{bommes2013quad}
D.~Bommes, B.~L{\'e}vy, N.~Pietroni, E.~Puppo, C.~Silva, M.~Tarini, and
  D.~Zorin, ``Quad-mesh generation and processing: A survey,'' in {\em Computer
  Graphics Forum}, vol.~32, pp.~51--76, Wiley Online Library, 2013.

\bibitem{hatcher2005algebraic}
A.~Hatcher, {\em Algebraic Topology}.
\newblock 2005.

\bibitem{hansen2019toward}
J.~Hansen and R.~Ghrist, ``Toward a spectral theory of cellular sheaves,'' {\em
  Journal of Applied and Computational Topology}, vol.~3, no.~4, pp.~315--358,
  2019.

\bibitem{lim2020hodge}
L.-H. Lim, ``{Hodge Laplacians on Graphs},'' {\em SIAM Review}, vol.~62, no.~3,
  pp.~685--715, 2020.

\bibitem{jia2019ssl}
J.~Jia, M.~T. Schaub, S.~Segarra, and A.~R. Benson, ``Graph-based
  semi-supervised \& active learning for edge flows,'' in {\em SIGKDD
  Conference on Knowledge Discovery and Data Mining}, pp.~761--771, 2019.

\bibitem{leblanc1975algorithm}
L.~J. Leblanc, ``An algorithm for the discrete network design problem,'' {\em
  Transportation Science}, vol.~9, no.~3, pp.~183--199, 1975.

\bibitem{hajijcell}
M.~Hajij, K.~Istvan, and G.~Zamzmi, ``Cell complex neural networks,'' {\em
  NeurIPS 2020 Workshop on TDA and Beyond}, 2020.

\bibitem{krizhevsky2012imagenet}
A.~Krizhevsky, I.~Sutskever, and G.~E. Hinton, ``{ImageNet} classification with
  deep convolutional neural networks,'' in {\em Advances in Neural Information
  Processing Systems}, pp.~1097--1105, 2012.

\bibitem{kipf2016semi}
T.~N. Kipf and M.~Welling, ``Semi-supervised classification with graph
  convolutional networks,'' {\em International Conference on Learning
  Representations}, 2017.

\bibitem{roddenberry2019hodgenet}
T.~M. Roddenberry and S.~Segarra, ``{HodgeNet}: Graph neural networks for edge
  data,'' in {\em Asilomar Conference on Signals, Systems, and Computers},
  pp.~220--224, 2019.

\end{thebibliography}

\end{document}